\documentclass[acmsmall, nonacm, screen, authorversion]{acmart}

\usepackage{amsmath}
\usepackage{subfig}
\usepackage{bbm}
\usepackage[ruled]{algorithm2e}
\usepackage{wrapfig}
\usepackage{booktabs}
\DeclareMathOperator*{\argmax}{arg\,max}

\begin{document}

\title{Consistency Training of Multi-exit Architectures for Sensor Data}

\author{Aaqib Saeed}
\affiliation{
  \institution{Eindhoven University of Technology}
  \city{Eindhoven}
  \country{The Netherlands}
}
\email{aqibsaeed@protonmail.com}

\renewcommand\shortauthors{A. Saeed et al.}

\begin{abstract}
Deep neural networks have become larger over the years with increasing demand of computational resources for inference; incurring exacerbate costs and leaving little room for deployment on devices with limited battery and other resources for real-time applications. The multi-exit architectures are type of deep neural network that are interleaved with several output (or exit) layers at varying depths of the model. They provide a sound approach for improving computational time and energy utilization of running a model through producing predictions from early exits. In this work, we present a novel and architecture-agnostic approach for robust training of multi-exit architectures termed \textit{consistent exit training}. The crux of the method lies in a consistency-based objective to enforce prediction invariance over clean and perturbed inputs. We leverage weak supervision to align model output with consistency training and jointly optimize dual-losses in a multi-task learning fashion over the exits in a network. Our technique enables exit layers to generalize better  when confronted with increasing uncertainty, hence, resulting in superior quality-efficiency trade-offs. We demonstrate through extensive evaluation on challenging learning tasks involving sensor data that our approach allows examples to exit earlier with better detection rate and without executing all the layers in a deep model. 
\end{abstract}



\maketitle

\section{Introduction}
Deep learning has achieved tremendous success in solving challenging problems in several domains, from audio recognition, healthcare, pervasive sensing, game-playing, to object detection, and more. Over the years, neural networks have progressively become better, wider, and deeper through improved understanding of the algorithms, development of techniques that introduce stronger inductive biases, and superior optimization methods that enable exceptional recognition capability. These advancements have been made possible due to the availability of better hardware (e.g., GPUs and TUPs) for learning and inference as well as massive labeled datasets. These modern techniques significantly improved the (which to some extent can be seen as one-time cost) learning phase of deep models. However, a large model size notoriously increases running cost as it is a continuous expense in production, hence it needs improvements as well.
In particular, it limits the adoption of deep models on resource-constrained devices like wearables with limited battery and computational power.

In addition, the huge latencies for inference present a significant challenge for deploying deep networks for real-time applications. For several applications in a real-world setting, the execution rate of the model is as important as the correctness of the obtained predictions. Thus, there has been increasing research interest in developing methods for accelerating neural network execution, especially for improving running-time and energy efficiency trading it against quality of the output~\cite{rastegari2016xnor,luo2017thinet,gupta2015deep,lane2018deep,jacob2018quantization}. Similarly, studies on improving the understanding of learning in neural networks have shown that the smaller network architecture with fewer layers can reliably identify patterns of interest in challenging datasets. For instance,~\cite{Leroux2017TheCN} showed that $33\%$ top-5 accuracy could be achieved on ImageNet, while solely using features from the first convolutional layer. Similarly,~\cite{kaya2019shallow} demonstrated an \textit{over-thinking} phenomena in neural networks that the predictions based on features from earlier layers are correct but become incorrect with progressively deeper architectures, resulting in wasteful computation.   %

Traditionally, the computational requirements of the model are decided during design time, which result in a model with fixed output accuracy. The appropriate architecture is either precisely designed or post hoc methods, e.g., quantization and pruning are applied to generate an optimized model that meets the inference budget for particular hardware. However, this strategy can lead to sub-optimal results when the budget and deployment hardware are not known in advance, which is the case in several practical settings. A fixed architecture cannot completely utilize the system capabilities in a dynamic environment, i.e., if additional computational power is available. Likewise, a highly accurate (but non-optimized) model might fail to produce satisfactory results under limited allocated resources. Even more importantly, large-scale well-annotated data are required to generate a high-performing model, which is prohibitively expensive to accumulate. To overcome these limitations, we present an approach to accelerate neural network inference through leveraging a multi-exit architecture design and develop a methodology to improve the performance of the exits through leveraging weak supervision as there is limited annotation opportunity in several domains, e.g., ambient intelligence. The early-exit models are capable of trading-off computation and accuracy dynamically on a per-instance basis for efficient utilization of the available resources~\cite{huang2017multi,teerapittayanon2016branchynet,kaya2019shallow,laskaridis2021adaptive} based on entropy of the network output or other criteria. The improvement in inference cost is achieved through having multiple exits or intermediate decision layers in a deep network at different depths and selecting an appropriate exit as needed based on a predetermined threshold.

We present a consistent exit training (CET) framework for training multi-exit architectures, where, \textit{consistent} derives from the usage of consistency objective in learning phase of the network to make it robust to input perturbations. With CET, we leverage weak supervision to produce a model with better generalziation through aligning model output on clean and noisy examples. Conventionally, a multi-task learning formulation is adopted for training a network with many exits, where each exit is considered a task classifier and it has a separate loss (or objective) function. The losses from all given exits are aggregated, and a sum of these is minimized with a gradient-based method. Although this naive formulation aims to produce exits with reasonable performance, it ignores prior knowledge about the neural network architecture. For instance, the earlier decision layers have lower capacity and will be less accurate than their later counterparts \cite{phuong2019distillation}. Given this observation, we propose a consistency-based objective that aims to improve the performance of all the exits in the network. The key idea behind our consistency learning strategy~\cite{sajjadi2016regularization,miyato2018virtual,sohn2020fixmatch} is to make the model's exits robust through training it to produce similar predictions on clean as well as a perturbed version of the inputs.

Our technique induces a regularization through enforcing prediction invariance over different input perturbations. The consistency loss does not require access to the actual ground-truth and can be applied to the network predictions of clean and perturbed inputs. When consistency loss is combined with a standard loss function, such as cross-entropy for the classification task, the joint-loss optimization significantly improves the performance compared to a naive multi-exit loss. It also provides a systematic way for unsupervised data augmentation or incorporation of unlabeled data for semi-supervised learning though we did not explore it further in this work and leave its thorough study for the future work. Furthermore, as compared to the distillation-based auxiliary loss~\cite{hinton2015distilling,phuong2019distillation}, our technique can improve the performance of all the exits, including the last, which is not possible with the former method as the last exit acts as a teacher for providing supervision to the earlier exits. Nevertheless, our method can be extended in a straightforward way to incorporate knowledge distillation (see Section~\ref{sec:results}). Likewise, our approach is orthogonal to compression and other strategies for improving neural network efficiency. It can be easily combined to further enhance the computational run-time or energy utilization of the neural networks.

This paper makes the following contributions:
\begin{itemize}
    \item We propose a novel \textit{consistent exit training} framework for learning multi-exit architecture with weak supervision. Our simple and architecture-agnostic approach exploit consistency regularization to enforce prediction invariance over clean and noisy inputs to improve the quality of exits under varying degrees of output uncertainty.
    \item We demonstrate the effectiveness of our method against several baselines on classification tasks involving sensor data that is largely generated by resource constrained devices. Our technique significantly improves the predictive performance of multi-exit architectures for various quality-efficiency trade-offs.
    \item We perform ablation to understand various design choices for input perturbation and ground-truth label generation schemes for consistency objective.
    \item Our framework can also be used to incorporate unlabeled data in the training procedure with no effort for semi-supervised learning of sensing models.
\end{itemize}

\noindent The rest of the paper is structured as follows. In Section~\ref{sec:background}, we provide an overview of the background and related work. Section~\ref{sec:method} presents our proposed consistent exit learning framework and other important related details, Section~\ref{sec:experiments} discusses an evaluation setup, datasets, and key experimental results to highlight the efficacy of our method for training multi-exit models. Finally, Section~\ref{sec:conclusion} concludes the paper and lists interesting directions for future research.

\section{Background and Related Work}
\label{sec:background}
To improve inference cost, efficient deep learning techniques gained significant attention due to their impact on embedded devices with limited memory, energy, and computational bandwidth~\cite{lane2018deep,xu2015cost, molchanov2016pruning}. In this section, we provide background on several related techniques and cover prior work before describing our CET framework. 

\subsection{Knowledge Distillation}
Distillation is a general-purpose technique for transferring knowledge between models to achieve smaller model with smaller size and less computational load~\cite{hinton2015distilling,bucilua2006model}. It aims to capture and transfer information from a large model (or ensemble), termed as a teacher for supervising a relatively small model named a student with the goal of achieving compression for efficient inference. In knowledge distillation, the teacher is generally a fixed pretrained network learned with a massive amount of high-quality data. In contrast, the student is a low capacity network that is guided to mimic the output of the teacher on a certain task. The rich supervisory signal from the privileged teacher model enables a compact student network to learn important aspects of the input that otherwise could not be possible while solely minimizing a supervised objective. Hinton et al.~\cite{hinton2015distilling} proposed to use softened class probabilities from the teacher to provide extra supervision, which acts as targets for the student model to optimize.

In the classical distillation framework, the teacher network $f_T$ produces logits $z_T$ (or unnormalized probabilities), which are a pre-activation vector before applying the softmax function ($\sigma$). In the case of ensembles or having multiple teachers, their outputs can simply be averaged. Similarly, a student network $f_S$, which has possibly different architecture and set of weights, produces an output analogous to the teacher model. The student network is trained to match its output as closely as possible to the teacher on the same input together, along with minimizing cross-entropy loss on the actual class labels $y$. The loss applied to the logits from $f_T$ and $f_S$ is temperature-scaled with a hyper-parameter $\tau > 1$ to soften the predictions. This step is crucial to mitigate the over-confidence of the networks as, generally, they put excessive probability mass on the top predicted class and too little on the rest. Thus, the scaling operation produces a more informative training signal from the teacher network as the difference between the largest and smallest output values is reduced. The complete learning objective that is minimized to penalize the difference between these models can be specified as:

\begin{align}
    \mathcal{L}_{Distillation} = \sum_{x_{i} \in X} l[f_T(x_{i}), f_S(x_{i})], \hspace{0.2cm} l = -\tau^{2} \sum_{k = 1}^{K} \big[  t_{k}  \cdot  \log(s_{k}) \big], 
    t(x) = \sigma{ \big( \frac{f_{T}(x)}{\tau} \big)},
    s(x) = \sigma{ \big( \frac{f_{S}(x)}{\tau} \big) }
\end{align}

Due to its conceptual simplicity, distillation is also adopted for improving the training of multi-exit architectures~\cite{phuong2019distillation} through treating later (generally last) exits in the network as a teacher and earlier exits as students. Nevertheless, due to dependence on the softmax function (or the number of classes) and differences in the model capacities, it is found to be difficult to incorporate the logits information from a teacher into a low-capacity student~\cite{cho2019efficacy, wang2020knowledge}. Therefore, distillation may also affect the learning process of multi-exit architectures, where early exits rely on fewer layers to produce the output. Conversely, our proposed approach aims to directly improve each decision layer's performance by enforcing output consistency without requiring a teacher model. Hence, this enables us to improve the last exit's recognition capability, which is not possible with distillation-based losses. Notably, our formulation can be extended in a straightforward manner to include a distillation objective if required, although we found it redundant in experiments.

\subsection{Consistency Training}
The aim of consistency training (or regularization) is to enable the model to be invariant to the noise in either the input or in the latent domains as a generalizable model should be robust to small deviations~\cite{miyato2018virtual,sajjadi2016regularization, xie2019unsupervised,Berthelot2019MixMatchAH}. Formally, given a learned model $f_{\theta}(.)$, an input $x$, and a noisy version $\hat{x}$ of an existing input, the network should produce the same output $y$ for both instances. The consistency loss helps achieve this goal by explicitly enforcing prediction invariance during the learning stage of the model by optimizing the following objective: 

\begin{align}
    \label{eq:const_loss}
    \mathcal{L}_{Consistency} = \mathbb{E}_{x \in X} \mathbb{E}_{\hat{x} \sim t(x)} \big[ \mathbbm{H}(f_{\tilde{\theta}}(y \mid x) ,f_{\theta}(y \mid \hat{x}) )\big]
\end{align}

\noindent where $X$ represents a dataset, $t(.)$ is a perturbation function, $\mathbbm{H}$ denotes a standard cross-entropy loss, $\tilde{\theta}$ shows a fixed network's weights same as parameters $\theta$ with the only difference that gradients are not propagated through it.

Due to the intuitive property of a self-training network without requiring labels, the consistency framework combined with pseudo-labeling is widely adopted for semi-supervised learning as it can effectively leverage large amounts of unlabeled data. The various methods in this area generally differ in the procedure of noise injection that varies from dropout, gaussian, or adversarial noise. Recently, the augmentation techniques, which are generally used for supervised training of deep models, are adopted as a form of strong corruption or perturbations. The methods based on augmentation have shown greater performance improvements compared to their weaker counterparts~\cite{xie2019unsupervised,Berthelot2019MixMatchAH}. Along these lines, the most relevant work is UDA~\cite{xie2019unsupervised} and FixMatch~\cite{sohn2020fixmatch} which proposed to use consistency in addition to standard classification loss for unsupervised data augmentation and semi-supervised learning, respectively. In particular, the FixMatch applies consistency loss over weakly and strongly augmented versions of the same image via treating predictions on weaker versions as pseudo-labels to improve the performance of an image classification model. Similarly, the detection of transformed or augmented input is also found to be a successful pretext task in unsupervised learning~\cite{saeed2019multi,sarkar2020self}. Inspired by the success of these methods, we propose to extensively use signal transformations for input perturbations and consistency loss to improve the robustness of early-exits. To the best of our knowledge, our work is the first to show that the consistency objective can be effectively used for training multi-exit architectures even if the annotated data is limited.

\subsection{Pruning and Quantization}
In addition to knowledge distillation for model compression, pruning, and quantization methods are also widely studied for accelerating neural networks~\cite{lecun1989optimal,courbariaux2014training,jacob2018quantization,gupta2015deep,molchanov2016pruning}. The network pruning approaches remove redundant parameters, neurons, or filters that have the lowest effect on performance. The earlier works on this topic utilize the Hessian of loss~\cite{lecun1989optimal}, l$2$-norm of the weights~\cite{han2015learning}, tensor decomposition~\cite{kim2015compression} and particle filtering based on misclassification rate to remove connections or parameters in a network~\cite{anwar2017structured}. Other methods increasingly focused on pruning and retraining strategies, i.e., removing nodes or channels in convolutional networks and retraining to compensate for the drop in performance~\cite{molchanov2016pruning,luo2017thinet}. More sophisticated techniques are concerned with designing efficient models~\cite{iandola2016squeezenet} and finding \textit{winning-tickets} i.e, sub-networks within a larger network that performs equally well~\cite{frankle2018lottery}. However, as shown in~\cite{hooker2019compressed}, the sparsity introduced through pruning can have consequences for certain types of examples in the dataset, and different classes can be impacted disproportionally. Likewise, network quantization is explored in a similar spirit to pruning, but instead of removing the model's parameters, it reduces their numerical precision and therefore speeds up the execution of floating-point operations and decreases the model size~\cite{jacob2018quantization,gupta2015deep}. We note that these methods are complementary to our approach of improving network efficiency with multiple exits and can be easily applied for further reducing the model size and improving computational cost. Moreover, our method permits controlling the accuracy and speed trade-off through making decisions on early-exits, which is related to the dynamic control of layers with dropout in transformer models used for language modeling tasks~\cite{fan2019reducing}. Finally, our approach only requires a single-pass training in contrast to some earlier mentioned approaches that depend heavily on retraining the network. %

\subsection{Multi-exit Architectures}
\begin{figure}[t]
\centering
\includegraphics[width=\textwidth]{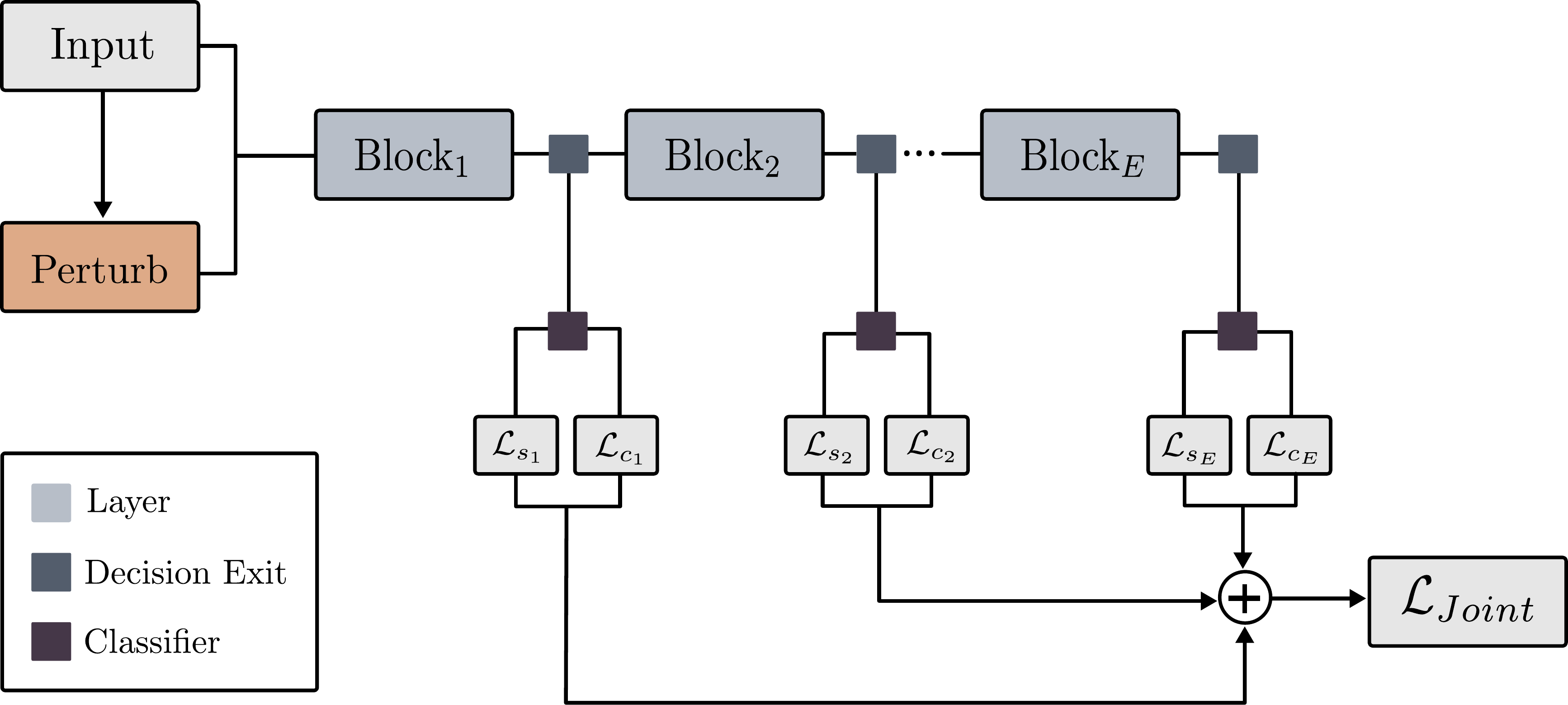}
\caption{Illustration of multi-exit architecture with consistent exit training. A deep model with multiple exits is trained jointly with consistency and standard classification objectives to become robust to input perturbations for efficient inference.}
\label{fig:overview}
\end{figure}

The multi-exit architecture (MEAs) specification enables us to build resource-efficient deep models through placing multiple exits at varying depths within a standard neural network~\cite{huang2017multi,kim2018doubly,teerapittayanon2016branchynet}, see Figure~\ref{fig:overview} for an illustration. Simply, stated the network based on MEA can be seen as a larger model comprising of numerous smaller sub-models with $E$ exits producing a list of outputs ($p_{e^1}$, $p_{e^2}$, ... $p_{e^E}$) for each of the decision exits. The MEAs are appealing for improving inference efficiency due to the observation that several instances are simple enough to not to be processed through an entire deep network~\cite{scardapane2020should,huang2017multi,kaya2019shallow}. For producing outputs (or labels) for easy to recognize samples, decisions could be made through executing a minimal number of layers, while complex examples are instead processed further by all layers in the network.  Hence, the MEA models are capable of trading off computation and accuracy dynamically on a per-instance basis through executing fewer layers for efficient utilization of resources ~\cite{huang2017multi,teerapittayanon2016branchynet,kaya2019shallow,scardapane2020should}. Specifically, at inference time, an instance passes through a layer (or block of layers) as usual; it is then forwarded to a decision layer before being processed in the later feature extraction layers. If an exit generates an output with sufficient confidence based on a predetermined threshold, the prediction is returned; otherwise, the input is processed in a similar fashion until the end. This strategy is termed “early-exit,” i.e., for a sample, an output can be produced at any time depending on the budget or accuracy.

Lately, several methods are proposed for MEAs or more generally early-exit inference from specialized neural network architectures~\cite{huang2017multi} to distillation-based training schemes~\cite{phuong2019distillation} and dynamic exiting through skipping layers~\cite{trapeznikov2013supervised,bolukbasi2017adaptive}, early-exit policy learning~\cite{wang2018skipnet} but with a concentration solely on vision and language modeling domains. Here, we provide a model-agnostic method for improving the recognition capability of early exits in the network with limited labeled data for learning. Compared to earlier work, we do not utilize a specialized architecture, and our technique can improve the performance of later exits too, as it does not rely on the teacher-student training paradigm. The simplicity of our approach also allows it to be easily combined with other methods to further advance the performance, for instance, through policy learning for early exit~\cite{wang2018skipnet}. Orthogonal to this line of work, networks with multiple auxiliary classifiers are also used in deep neural networks to mitigate the issue of vanishing gradients and avoiding overfitting~\cite{lee2015deeply,szegedy2015going}. For a detailed treatment of the topic, an interested reader is referred to an excellent review by Scardapane et al.~\cite{scardapane2020should}.

\section{Method}
\label{sec:method}
\subsection{Overview}
The multi-exit architectures (MEAs) are standard neural networks, comprising of several layers or sub-modules. Their key distinguishing factor is that instead of a single decision or exit layer at the end, they are augmented with several exits, which are placed at different depths of the network; see Figure~\ref{fig:overview} for an illustration. For a MEA with $E$ exits, the network has multiple decision layers ($f_{\theta}^{1}, f_{\theta}^{2}, \ldots, f_{\theta}^{E}$), that map an input $x$ to an output $y$ which is a probability complex for classification problems and a continuous outcome for regression-based tasks. It is important to note that, for the exits $1 < i <= E$, an exit $e^i$ is more accurate and more expensive to compute than the previous exit i.e., $e^{i-1}$. For accelerating inference, a slightly less accurate exit can be used for predictions based on a pre-defined threshold of the output, e.g., entropy. The motivation for this comes from observations in the vision and language community, where it has been shown that the higher layers in deep models learn fine-grained features~\cite{lecun2015deep}, but simple instances can be recognized reliably through coarse representations from earlier layers~\cite{kaya2019shallow}. Therefore, it may suffice to utilize intermediate layers for inference to achieve a trade-off between accuracy and computational effort. In principle, for attaining the balance, the MEAs are required to have a reasonable detection rate for all its exits, so an appropriate decision layer can be selected at runtime. Nevertheless, the later exits in a network generally utilize the computation of earlier layers and can also share the weights of the decision layers to better utilize resources.

The objective of our work is to improve the performance of MEAs, particularly of the earlier exits that have a low capacity to learn generalizable representations but are efficient in execution. Through enhancing the quality of decision exits, we aim to reduce the cost of inference that permits the execution of fewer layers while maintaining an optimal accuracy. Therefore, the model can generate predictions for as many samples as possible while adhering to the availability of system resources in a dynamic environment. To this end, we propose a framework named \textit{consistent exit training} for improved training of multi-exit architectures. It utilizes a novel objective function for enforcing prediction invariance over original and perturbed (or noisy) versions of the inputs, in addition to optimizing the task-specific loss. Specifically, in this work, we focus on classification-based tasks, but the framework can be used in a straight-forward manner for regression problems. Our approach enables the network to be more robust to small changes in the input through learning better features and aligning the decision layers to produce similar output regardless of the input perturbations. Furthermore, a distinctive property of our technique is that it allows seamless integration of unlabeled data in the learning process as it does not depend on the availability of annotated inputs as loss is applied on the network outputs on clean and perturbed inputs. In the following subsections, we provide key details of the framework, procedure of noise generation, achieving semi-supervised learning, architecture specification, and other training details. 

\subsection{Consistent Exit Training}
\label{sec:cet}
In \textit{consistent exit training} (CET), we consider a dataset $\mathcal{D} = \{(x_m, y_m)\}_{m=1}^{M}$ with $M$ instances, where, $x$ and $y$ represent the raw inputs and class labels, respectively. To improve the recognition quality of exits in a MEA model, we propose a novel training scheme based exclusively on two loss terms, a consistency objective $\mathcal{L}_c$ with relative weight $\lambda$ and a task-specific loss $\mathcal{L}_{s}$ (i.e., a cross-entropy loss function) that are jointly optimized for learning to solve a desired task as $\frac{1}{M} \sum_{m=1}^{M} \big[ \mathcal{L}_{s} + \lambda \mathcal{L}_{c} \big]$. For the consistency objective, given an input $x$, we compute the output distribution with a probabilistic model $p = f_{\theta} (y \vert x)$ and of a perturbed version $\hat{p} = f_{\theta} (y \vert t(x))$, where, $t(\cdot)$ represents a transformation or perturbation applied on a given input. To minimize a consistency loss $\mathbbm{H}(f_{\theta} ( y \vert x), f_{\theta} (y \vert t(x)))$ to match the predicted class distribution $p$ made with the network on clean (or original) input to those generated from a noisy version $\hat{p}$, we need to create artificial labels for each instance which are then used in divergence measure $\mathbbm{H}$ e.g. cross-entropy or Kullback–Leibler divergence to provide weak supervision. For computing the labels, we utilize $p$ to extract the pseudo-label $\bar{y} = \argmax(p)$, then we enforce prediction invariance via cross-entropy against model's output $\hat{p}$ on the noisy input as:

\begin{align}
    \label{eq:c_loss}
    \mathcal{L}_c = \frac{1}{M} \sum_{m=1}^{M} \big[ \mathbbm{1}(\max (p_m) \geq \kappa) \cdot \mathbbm{H}(\bar{y}_m, f_{\theta}(y \vert t(x_m) )) \big]
\end{align}

\noindent where $\kappa$ represents a confidence threshold above which labels on clean inputs are retained as ground-truth and $\mathbbm{1}$ is an indicator function. We utilize a dynamic confidence thresholding based on cosine schedule to gradually increase the $\kappa$ from $50\%$ to $90\%$. The motivation for using dynamic scheme as opposed to fixed constant value is that, in the beginning of the training network is less certain about the predicted classes but becomes more certain as training progresses. Therefore, gradually selecting highly confident examples to provide weak supervision results in a better training signal and provides an opportunity to the network to learn via a form of curriculum learning (i.e., from easy examples first and then later from hard examples). It is important to note that equation~\ref{eq:c_loss} resembles a pseudo-labeling objective with a crucial difference is that it is being applied over perturbed inputs. Thus, the consistency loss induces a stronger form of regularization, which, as we show in section~\ref{sec:results} that significantly improves the model performance compared to other competitive baselines. Likewise, the classification loss is computed over model predictions on clean inputs and semantic labels $y$ in a standard way for training MEA network as:

\begin{align}
    \label{eq:xe_loss}
    \mathcal{L}_{s} = \frac{1}{M} \sum_{m=1}^{M} \big[ \mathbbm{H}(y_m ,f_{\theta}(y_m|x_{m})) \big]
\end{align}

\begin{algorithm}[t]
\caption{Consistent Exit Training}
\label{alg:cel}
\KwIn{Dataset $\mathcal{D}$ comprising $N$ instances, number of iterations $I$, batch size $B$, Input perturbation functions $T$}
\KwOut{Trained multi-exit model $f$}
initialize a multi-exit network $f$ with parameters $\theta_{f}$\\
\For{iteration $i$ $\in$ $\{$ $1$, \ $\ldots$, \ $I$ $\}$ }
{
    Randomly sample a mini-batch of $B$ instances from $\mathcal{D}$ as $\{$$(x$, $y)_1$, $(x$, $y)_2$, $\ldots$, and $(x$, $y)_b$$\}$ \\
    Generate perturbed inputs from each example by randomly selecting a $t \in T$ and add to a batch\\
    Update $\theta_f$ with gradient-based optimization with an aggregate of exit-wise and consistency losses over exits\\
    $\nabla_{\theta_{f}} \Big[\frac{1}{M} \sum_{m=1}^{M} \mathcal{L}_s(y_m,f_{\theta}(x_m)) + \lambda \frac{1}{M} \sum_{m=1}^{M} \mathcal{L}_c(\bar{y_m},f_{\theta}(\hat{x_m}))]$
}
\end{algorithm}

We provide a complete algorithm for consistency learning of multi-exit architectures in Algorithm~\ref{alg:cel}. We add a joint loss term to each exit $e$ in the network and train all the exits together through optimizing an aggregated loss $\frac{1}{E}\sum_{e \in E} \mathcal{L}_{e}$ with a gradient-based method. Importantly, for computing loss and gradient updates based on consistency loss, we treat the model's output $\bar{y}$ on clean input $x$ as constants in equation~\ref{eq:c_loss} to make sure that the gradients are not computed in an inconsistent way, and only the output of a clean unperturbed input can be used to teach the network and not the other way around. By means of consistency training, the network becomes robust to corruption by aligning its output with different types of input noise. In addition, our approach provides several other benefits apart from faster inference through early exits, including mitigation of vanishing gradients due to earlier decision layers, providing additional gradient signals,  stronger regularization with input perturbations, and joint objective optimization, which results in a highly generalizable model. Likewise, our method can utilize unannotated input in a straightforward way as a consistency loss controls the gradual propagation of label information from labeled samples to unlabeled ones for teaching the network.

\subsubsection*{\textbf{Input Perturbations}}
A key element in the consistency training, is the perturbation function to apply on the input or latent space representations, i.e., either input $x$ or some intermediate layer's output $z$. Here, we propose to use transformation in the input domain as it has been shown to significantly enhance the network performance in various areas~\cite{xie2019unsupervised,sohn2020fixmatch,saeed2019multi,um2017data,Tang2021selfhar}. Likewise, it also makes intuitive sense as we can visually inspect the signal in its original form to verify that the perturbed version does not look drastically different than the clean input and thus preserves the label. Conversely, the perturbations in latent space are difficult to achieve as the noise might cause the learned embedding (or features) to be far away from the manifold of those of the actual samples and hence lead to poor model generalization. However, we note that augmentation in the latent space is an open area of research~\cite{Devries2017DatasetAI,Li2020OnFN} and our method can reliably incorporate such strategies to impose consistency without any change to the underlying framework. Therefore, inspired from the success of the transformations in vision, text, and sensory domains~\cite{xie2019unsupervised,saeed2019multi}, we utilize the following perturbations to minimize the discrepancy between the network's outputs via a consistency objective:

\begin{itemize}
    \item{\textbf{Additive and Multiplicative Noise:} For noise injection, we uniformly sample a noise tensor $\Gamma \sim \mathcal{N}(\mu,\sigma)$ of the same size as $x$ then we create $\hat{x}$ as $x + \Gamma$ which adds random noise to each sample. For scaling the magnitude, we multiply the input samples with a randomly selected scalar for each channel dimension to produce $\hat{x}$. The noisy perturbations are widely used in one form or another to improve the robustness of neural networks. Similarly, the utilized transformations help the model to generalize better as it becomes invariant to corruption and amplitude shifts of the input, which may be caused by malfunctioning sensors and other heterogeneities of the system.} \\
    \item{\textbf{Time Warping:} It is used to apply local stretching or warping of a signal for smoothly distorting time intervals between samples with randomly generated cubic splines for each input channel. This transformation generates $\hat{x}$ that preserves the underlying characteristic of the input where an event of interest within a signal might be translated along the temporal dimension.} \\
    \item{\textbf{Masking:} To reduce the network's reliance on specific segments of input to correctly recognize its label, we utilize masking. A segment $s$ of length $l$ is randomly selected from a signal, and the values within that region are zero-out while leaving other samples unchanged.} \\
\end{itemize}

\noindent These transformations or augmentations provide additional benefits compared to merely using Gaussian noise, which makes local alterations to the input. The synthetic examples produced with the above-discussed functions are not only realistic but diverse. Likewise, the choice of our perturbations are more widely applicable to a range of signals without changing the underlying characteristic of the input and affecting the corresponding class label. Hence, imposing consistency between clean and perturbed inputs is reliable and regularizes the model sufficiently to become robust to a variety of noise types.

\begin{wrapfigure}{R}{0.48\textwidth}
    \begin{minipage}{0.48\textwidth}
        \begin{algorithm}[H]
        \caption{Efficient Inference via Early Exit}
        \label{alg:eei}
        \KwIn{Instance $x$, Entropy threshold $\varphi$}
        \KwOut{$y$}
        \For{$i \gets 1$ to $E$}
        {
            $\hat{y}^{i}$ =  $f_{\theta}^{i}(x)$\\
            $\epsilon$ = entropy($\hat{y}^{i}$)\\
            \uIf{$\epsilon$ < $\varphi$}{
                return $\hat{y}^{i}$
            }
        }
        return $\hat{y}^{E}$
        \end{algorithm}
    \end{minipage}
\end{wrapfigure}

\subsubsection*{\textbf{Faster Inference with Early Exit}} 
To improve the network acceleration for rapid inference, we estimate the decision layer's confidence for its prediction with entropy $\epsilon$ of the output distribution $y_m = f_{\theta}(x_m)$. The output is computed in a standard way through a forward-pass from a model with a non-perturbed input. When an instance $x_m$ reaches an exit $e$, its probability distribution's entropy is compared with a pre-specified threshold hyper-parameter $\varphi$. If the computed entropy $\epsilon$ value is within the defined boundary, the prediction is returned, and the inference stops; otherwise, the sample proceeds through the next set of layers following an identical approach till the end. This early-exit procedure is summarized in Algorithm~\ref{alg:eei}, and it is more beneficial when combined with our proposed strategy of training multi-exit models, which enables all the exits to be highly accurate and hence assure better resource utilization. Moreover, it is of significance to note that higher values of $\varphi$ lead to a faster inference but inaccurate predictions, and inversely smaller $\varphi$ induce accurate output but a slower inference. An optimal value of $\varphi$ can be set based on a performance on a development (or validation) set via hyper-parameter search.
  
\subsection{Network Design, Optimization, and Training Details}
We use convolutional neural networks (CNN) to implement multi-exit architectures with $1$D (i.e., temporal) convolutional layers. We use a network with the same configuration to highlight the robustness of our approach for the considered learning tasks. Our model consists of $5$ exits; a decision layer is placed after every block consisting of a convolutional, layer normalization and max pooling layers. In the beginning of model we add instance normalization layer to normalize the incoming input channel-wise. The convolutional layers have $8$, $16$, $24$, $32$, and $64$ feature maps each with a kernel size of $4$. The max-pooling is applied with a pooling size and strides of $4$ to reduce the time dimension. The output block has a similar structure across the network, which consists of a global average pooling to aggregate embedding. It is followed by a fully-connected layer with $32$ hidden units and a classification layer with units depending on the number of categories for a particular classification task. We apply layer normalization after convolutional layer and use \texttt{PReLU} activation in all layers except the last, L$2$ regularization is used with a rate of $1e-4$ and dropout with a factor of $0.1$ in second and fourth blocks after the max pooling layer.

We emphasize that a basic neural network can be extended in a straightforward manner through augmenting intermediate layers with decision exits to transform it into a multi-exit model that can be trained with our consistent exit learning framework. In our case, the choice of hyper-parameters and other architectural configurations is guided through early exploration of our approach on a subset of the training set (see section~\ref{sec:evaluation} for details about the datasets). For a single-exit baseline model, we use a same network architecture as multi-exit model with the only difference being that it contains decision exit layers at the end. Furthermore, for gradient-based parameter updates of the model, we use Adam optimizer with a learning rate of $0.0003$ for $100$ epochs with a batch size of $32$, unless mentioned otherwise. For consistency loss, we generate the perturbed input for each example within a batch by randomly selecting a function defined in section~\ref{sec:cet}. The pseudo-labels are created through $\argmax$ operation for instances on which the network has a \texttt{softmax} score (i.e., if a network is $80$\% confident that an example belongs to a particular class) above $\kappa$. The joint losses from each exit are aggregated to compute a total loss, where a consistency objective is weighted with a factor of $0.5$ to be minimized with gradient descent. For perturbed input generation, we randomly select an augmentation function for each example in the batch to create its noisy version. We use multiplicative and additive noise sampled from normal distribution with a standard deviation of $0.2$ and mean of $1.0$ and $0.0$, respectively. We perform masking with randomly selected sub-segments by zeroing out the values in an example, we use window length of $100$ and $500$ samples for HHAR and SleepEDF datasets, respectively (see Section~{sec:evaluation}). Likewise, for time-warping we generate random curves from a normal distribution with a standard deviation of $0.3$ to smoothly distort time steps.

\section{Experiments}
\label{sec:experiments}
\subsection{Evaluation Setup}
\label{sec:evaluation}
We conduct experiments on two sensory datasets for the tasks of activity detection and sleep stage scoring. We use Heterogeneity Human Activity Recognition (HHAR) dataset~\cite{stisen2015smart} to recognize activities of daily living from IMUs signals (i.e., accelerometer and gyroscope) collected from a smartphone. The data is collected with $8$ smartphones carried on different body parts of the subject.  The experimental setup of~\cite{stisen2015smart} used $36$ different smart devices (smartphones and smartwatches) of $13$ models from $4$ manufacturers to cover a broad range of devices for sampling rate heterogeneity analysis. In total, $9$ participants executed $6$ activities (i.e., biking, sitting, standing, walking, stairs-up, and stairs-down) for $5$ minutes to get equal class distribution. The sampling rate of signals varies considerably across devices, with values ranging between $50-200$Hz. We use signals collected from the smartphones in our analysis and segment them into fixed-size windows of $400$ samples with $50\%$ overlap and solely perform standard mean normalization of the input channels without any further pre-processing. For sleep stage scoring, we utilize The Physionet Sleep-EDF dataset~\cite{kemp2000analysis} consisting of $61$ polysomnograms collected from $20$ subjects to study the effect of a) age on sleep in healthy individuals and b) the effects of temazepam on sleep. The dataset includes $2$ whole-night sleep recordings of EEGs from FPz-Cz and Pz-Oz channels, EMG, EOG, and event markers. The signals are provided at a sampling rate of $100$Hz, and sleep experts annotated $30$ seconds segments into $8$ classes. The classes include Wake (W), Rapid Eye Movement (REM), N$1$, N$2$, N$3$, N$4$, Movement, and Unknown (not scored). We applied standard pre-processing to merge N$3$ and N$4$ stages into a single class following AASM\footnote{American Academy of Sleep Medicine} standards and removed the unscored and movement segments. We utilize the Fpz-Cz channel of EEG signal from the first study to classify sleep into $5$ classes, i.e., W, REM N$1$, N$2$, and N$3$.

We compare the performance of our consistency-based learning objective with several baselines: a) standard or single exit model, b) exit-wise loss, where losses from all exits are aggregated to update model parameters, c) a multi-exit network trained with augmentations and d) distillation loss, which employs knowledge distillation in addition to the exit-wise loss for multi-exit models. Specifically, it treats the last exit as a teacher whose knowledge is transferred to earlier exits that act as students. For performance evaluation, we divide the considered datasets into train/test splits of $70-30$ based on users with no overlap among these, i.e.; we keep $70\%$ of the users in training and the remaining $30\%$ in the testing set. We compute the accuracy, macro F1-score, and Cohen's kappa to be robust to the imbalanced nature of the datasets. In particular, it is important to note that the same network architecture is used for all considered learning tasks to highlight the effectiveness of CET.

\subsection{Key Results}
\label{sec:results}

\begin{figure}[t]
\centering
\subfloat[HHAR]{\includegraphics[width=.5\textwidth]{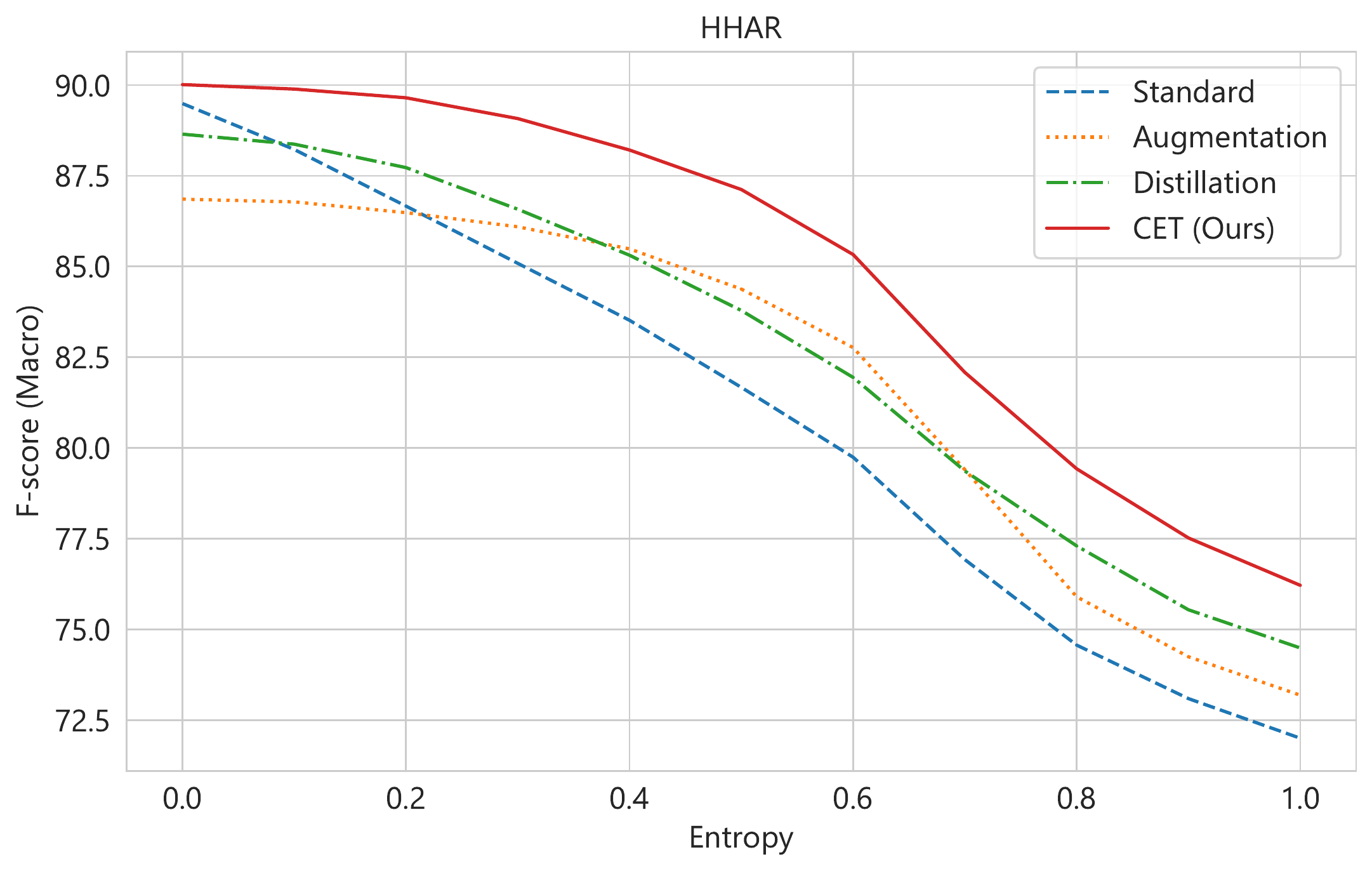}} 
\subfloat[SleepEDF]{\includegraphics[width=.5\textwidth]{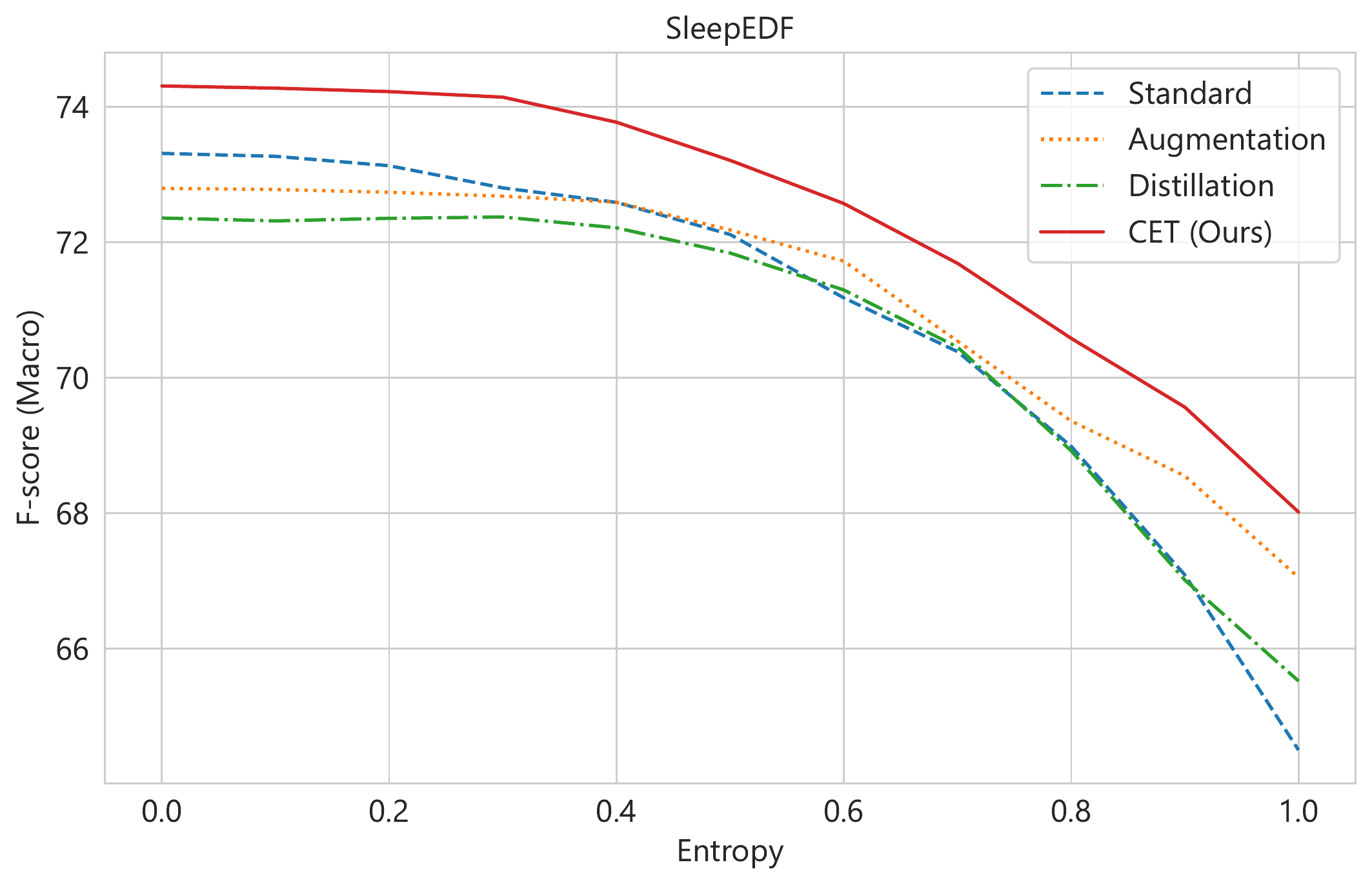}} 
\caption{Evaluation of our CET framework against several baselines for training multi-exit models. Our approach provides better recognition in comparison with other strategies and outperforms them significantly while maintaining a superior F-score when the entropy of the predictions increases and examples exit from the earlier exits.}
\label{fig:fscore_entropy}
\end{figure}

We compare the performance of our proposed consistent exit training framework with several baselines in Figure~\ref{fig:fscore_entropy}. Our approach significantly outperforms standard exit-wise training and provides better performance when the uncertainty of the predictions increases via gradually increasing the value of entropy from zero to one. The CET models also have better F-score as compared to distillation-based training, where last exits outputs are used as teacher supervision for earlier exits (students) to optimize an additional loss in addition to the exit-wise classification loss. In particular, we use a constant temperature scaling value of $2$ to soften the softmax outputs of the teacher to mitigate the model's overconfidence. Furthermore, we also compare against a multi-exit model trained with the same augmentations or input perturbations functions as used for CET to assess the effectiveness of our consistency objective. Although training with augmentations improves the performance over standard models, we notice that consistency training leads to better generalization on the considered learning tasks. Overall, our method is more robust to increasing uncertainty or via entropy values, as it enables a model to produce consistent outputs on a clean and perturbed version of the same example. Specifically, on HHAR, a multi-exit model trained with CET achieves an F-score of around $86\%$ as compared to $80\%$ of a standard model. Similarly, on the sleep stage scoring task, even when entropy is one, CET has an F-score of $68\%$ while the standard model scores around $64\%$. These results demonstrate the effectiveness of our consistency training approach in improving the generalization of multi-exit sensory models in a straightforward manner to achieve superior efficiency and recognition rate trade-off for devices with limited computational resources.

\begin{figure}[!htbp]
\centering
\subfloat[HHAR]{\includegraphics[width=.5\textwidth]{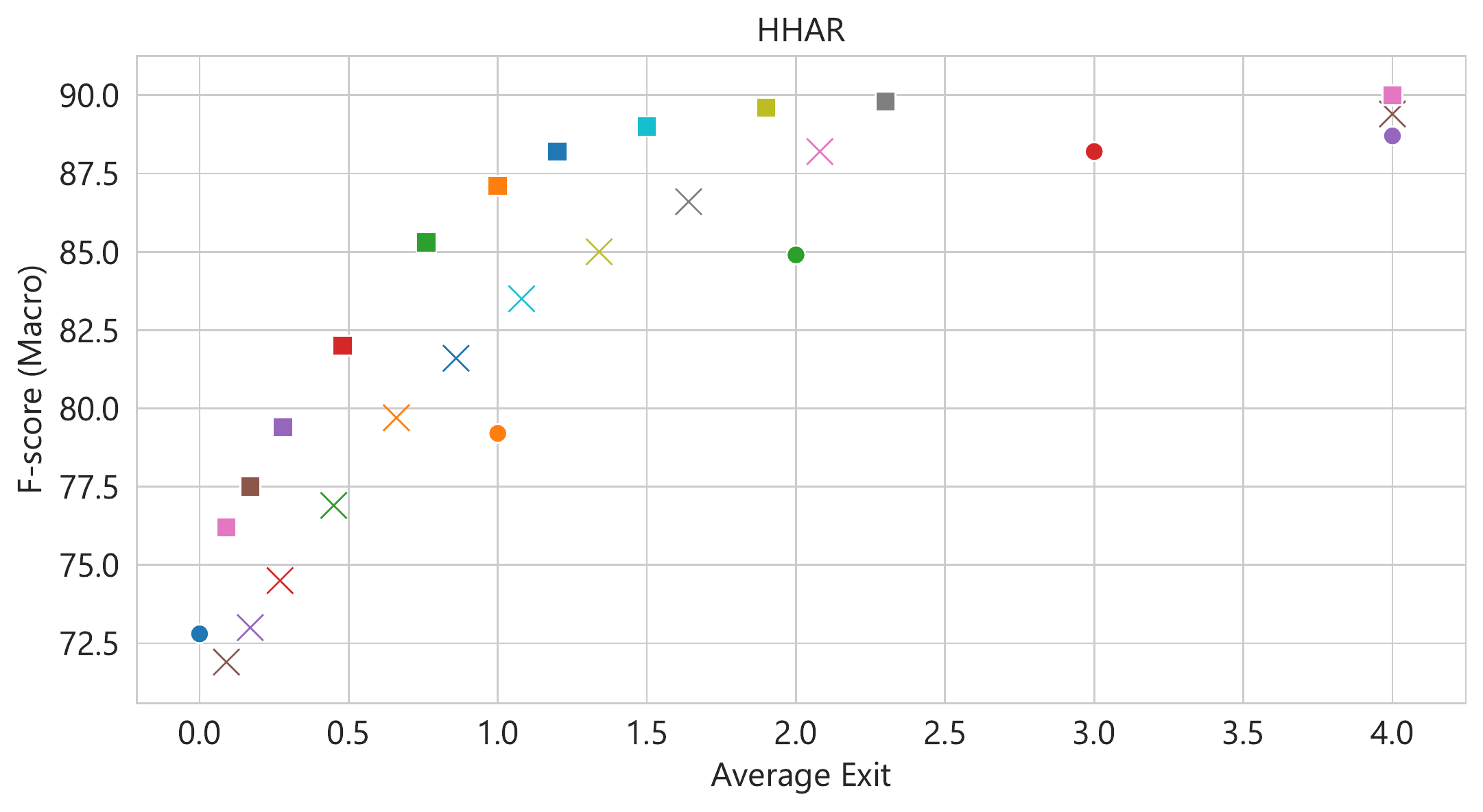}} 
\subfloat[SleepEDF]{\includegraphics[width=.5\textwidth]{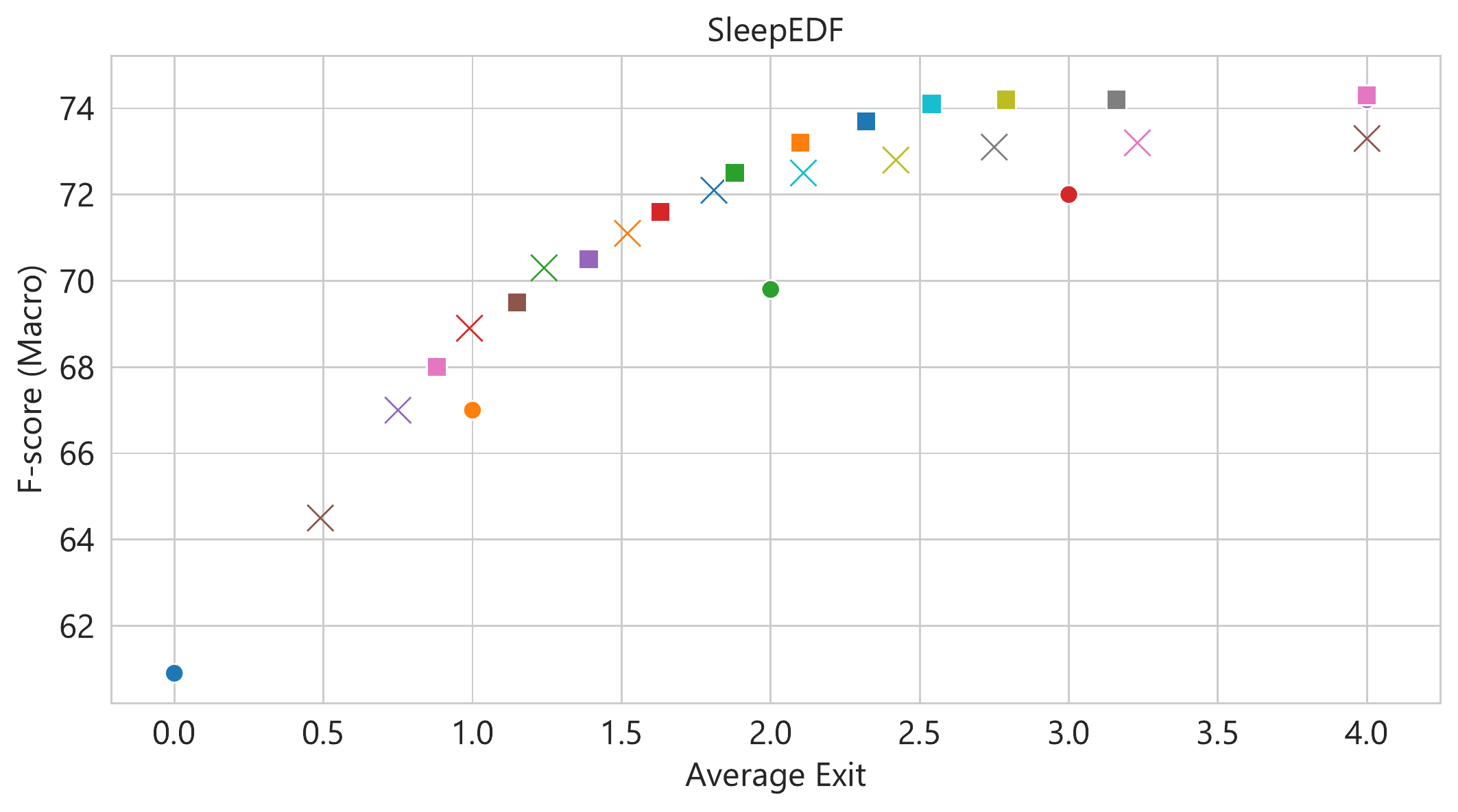}} \\
\subfloat{\includegraphics[width=.5\textwidth]{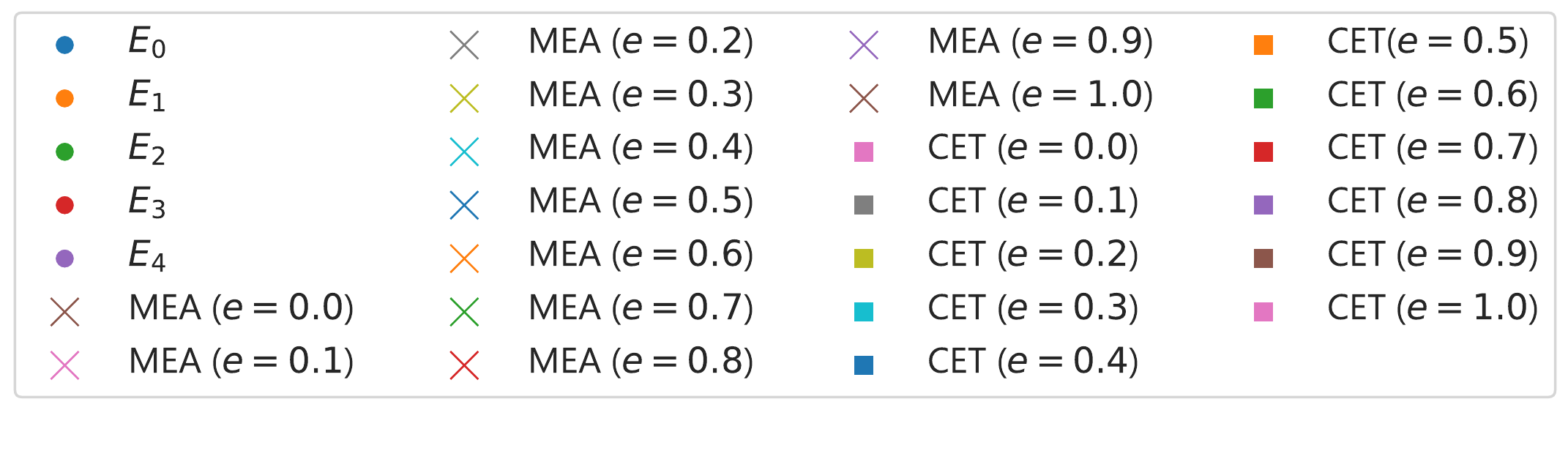}} 
\caption{Average exit and F-score trade-offs for a range of ($e$) entropy values. CET performs better than naive exit-wise loss for multi-exit models as well as training models with a single exit up to a defined number of layers.}
\label{fig:fscore_avgexit}
\end{figure}

In order to further understand the effect of consistency training, we study average exit and F-score achieved by multi-exit models trained with CET and exit-wise losses. Figure~\ref{fig:fscore_avgexit} provides F-score for various values of entropy and corresponding average exits, where the test set examples on average exit to satisfy the predefined entropy threshold. We also compare against training models with single-exit up to a specific number of layers, i.e., smaller sub-models from a multi-exit architecture. We notice in all the cases, multi-exit models trained with the consistency objective perform better than individual single-exit models, represented by $E_{k}$. Furthermore, CET models have better generalization compared to the naive multi-exit model in terms of F-score and average exit. In particular, we notice that in some cases, CET models prefer later exits (i.e., average exits are slightly higher for the same entropy threshold among MEA and CET) to improve the recognition rate as compared to naively exiting earlier, resulting in poor performance. These results hint that consistency training with input perturbation improves the model's capability to recognize hard-to-classify examples and hence defer such instances to later exits for better output quality.

\begin{figure}[!htbp]
\centering
\subfloat[HHAR]{\includegraphics[width=\textwidth]{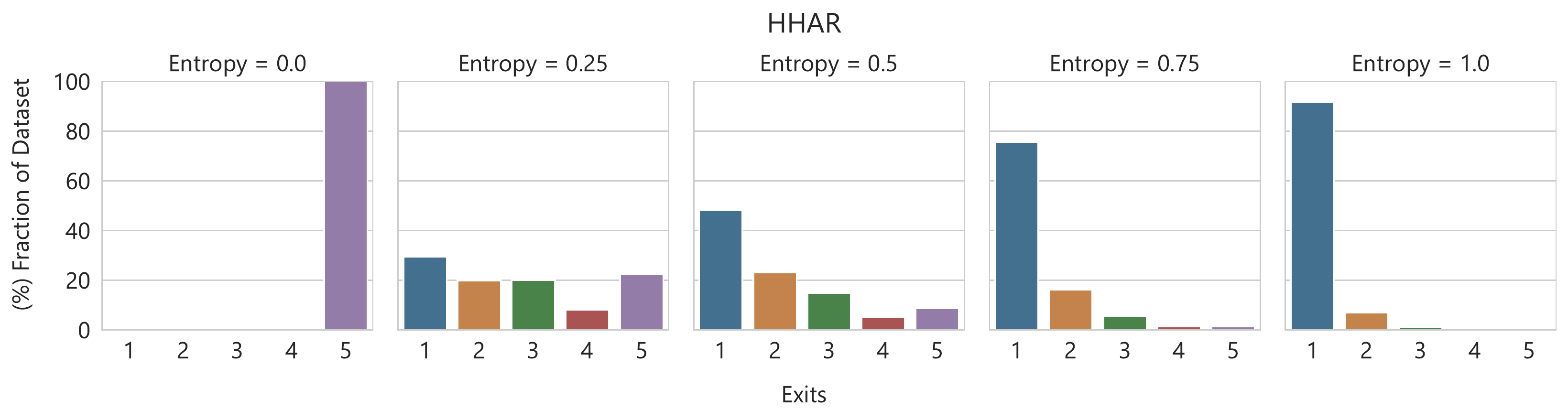}} \\
\subfloat[SleepEDF]{\includegraphics[width=\textwidth]{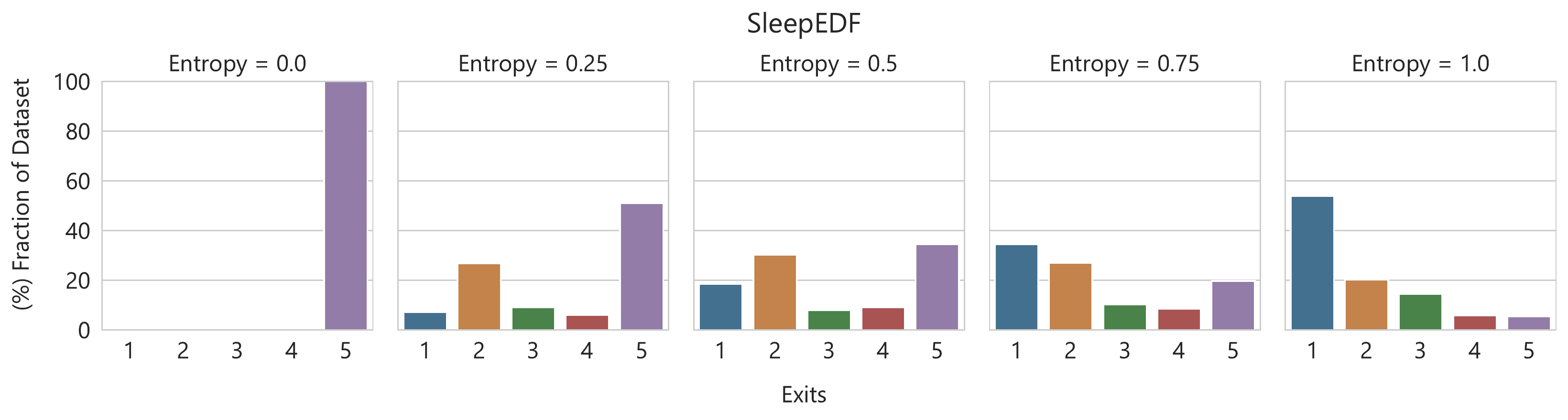}}
\caption{The number of output examples per exit for a model trained with our CET framework. Each plot represents a separate entropy threshold $\varphi$.}
\label{fig:fraction}
\end{figure}

We also demonstrate the fraction of samples exiting at each exit for a given entropy threshold in Figure~\ref{fig:fraction} on corresponding test sets for a model trained with CET. We treat entropy threshold $\varphi = 0$ as a baseline, where all examples leave at the last layer, but as $\varphi$ increases gradually, more examples exit earlier to leverage uncertainty in network output with a goal of improving usage of computational resources. Interestingly, we observe that on the human activity recognition task, a larger number of examples are reliably classified with earlier exits when $\varphi$ is greater than zero. In particular, for $\varphi > 0.75$ majority of the instances exits from the first exit but from Figure~\ref{fig:fscore_avgexit} we conclude that the entropy values between $0.3-0.5$ provide better F-score and average exit trade-off. Furthermore, on the SleepEDF dataset, we see that samples largely prefer second or last exits while few percentages of samples leverage other exits. Apart from the obvious, we observe an additional interesting pattern that the EEG samples in sleep stage scoring task even for a high value of entropy, i.e., $\varphi = 1.0$ utilize all exits on average as opposed to using solely earlier exits as in HHAR. It might be due to the hardness of examples and modality differences across considered datasets. Our results show that an entropy threshold combined with our CET framework is able to choose the fastest exit among those with comparable quality and single-exit models (see Figure~\ref{fig:fscore_avgexit}) and attain a good trade-off between quality and efficiency.

\begin{figure}[!htbp]
\centering
\subfloat[HHAR]{\includegraphics[width=0.5\textwidth]{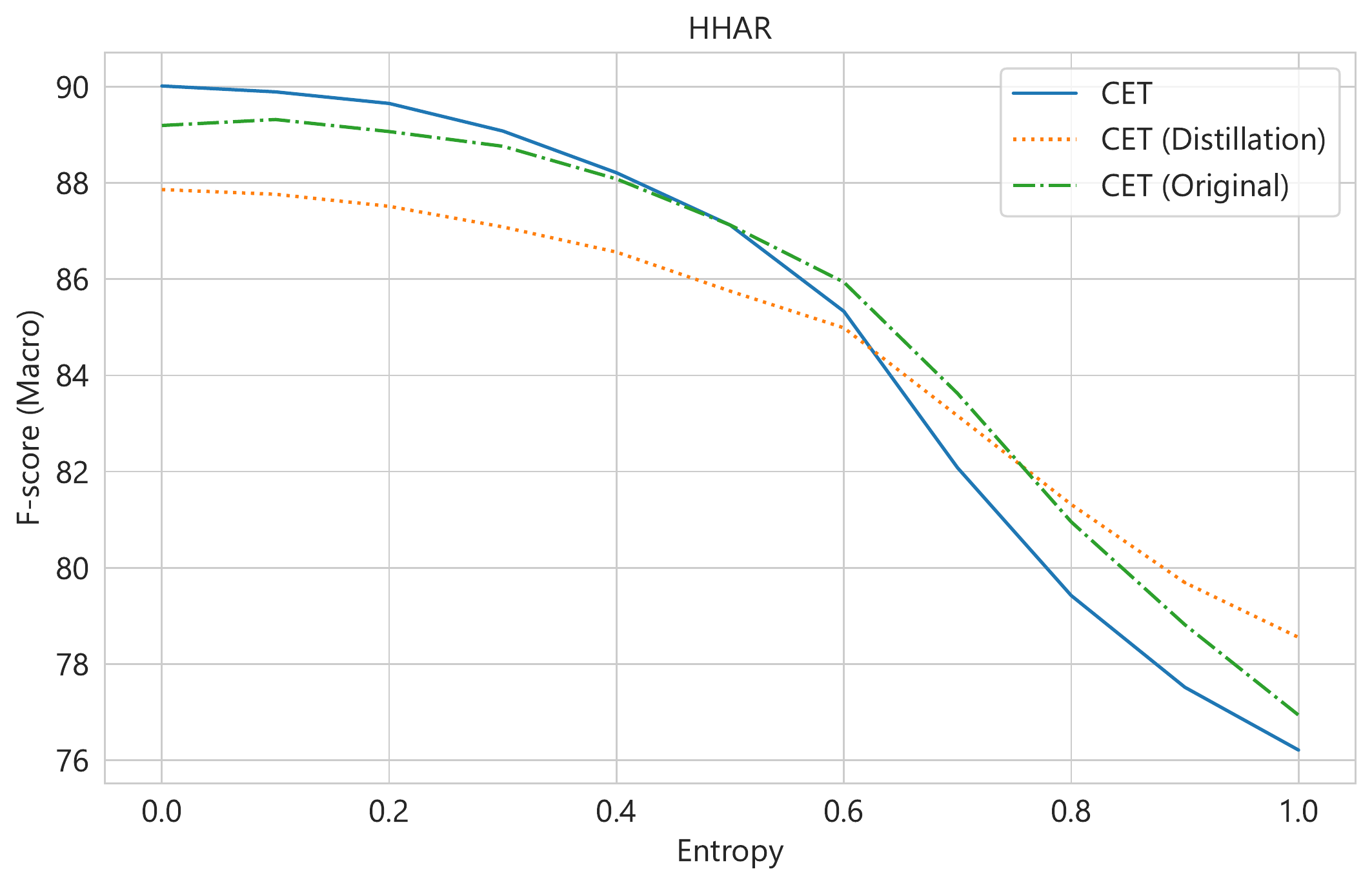}}
\subfloat[SleepEDF]{\includegraphics[width=0.5\textwidth]{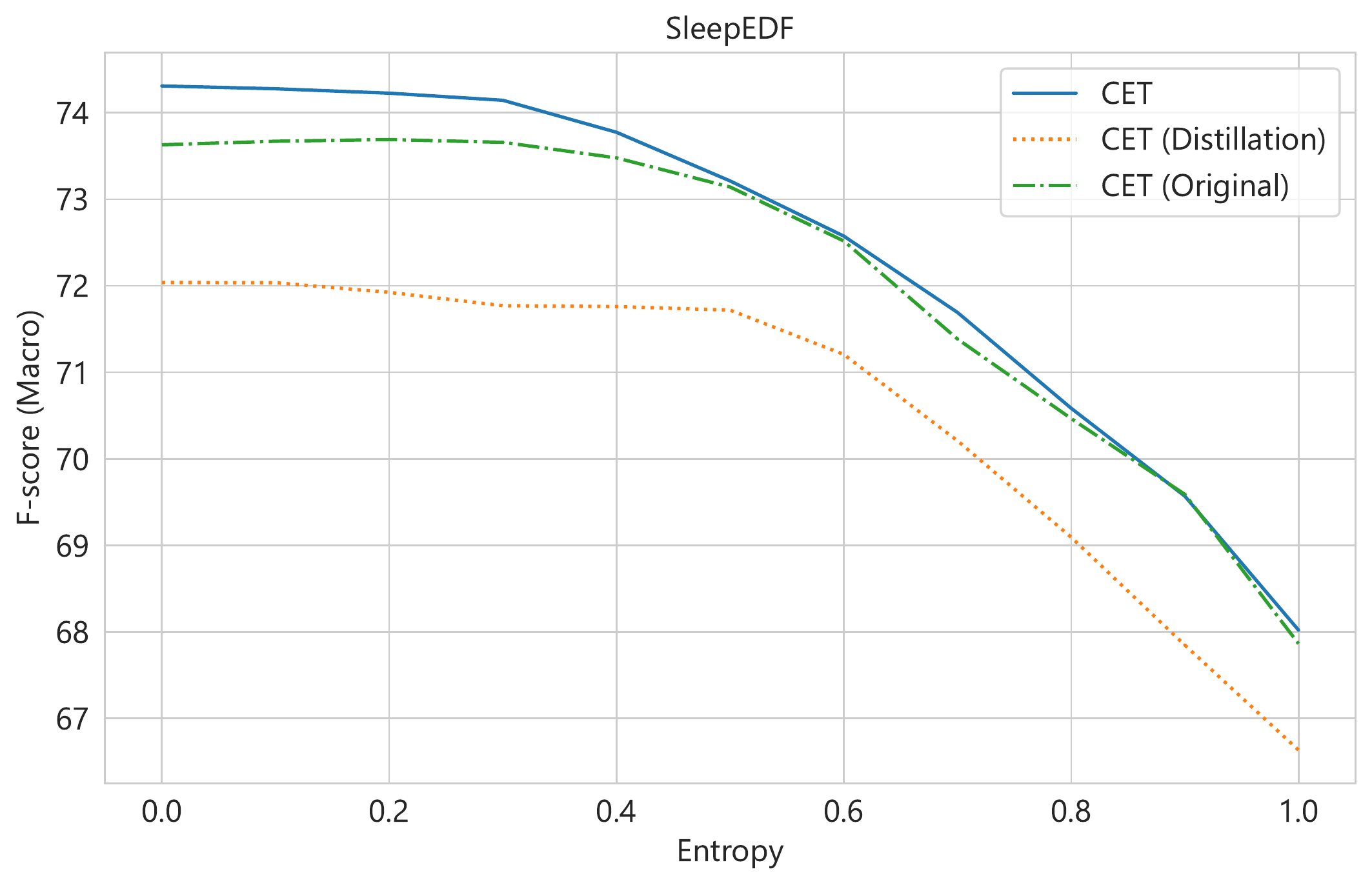}}
\caption{Assessment of consistent exit training with different forms of ground-truth label generation for computing consistency loss over perturbed inputs.}
\label{fig:cet}
\end{figure}

For all experiments reported so far, we used the consistency loss over each exit to enforce the network predictions on clean and perturbed inputs to be similar. We also perform exploratory studies on how the choice of ground-truth label generation can affect the overall performance. Figure~\ref{fig:cet} provides ablation results on these different choices of supervision for consistent exit training. In addition to the proposed approach as discussed in Section~\ref{sec:cet}, we explore distillation via the teacher, where the outputs of the last exit are treated as labels for the consistency loss for each exit. Similarly, as opposed to creating pseudo-labels via a teacher or from the same exits over clean examples, we use original class labels (e.g., activity classes) to compute consistency loss, the green dashed lines in Figure~\ref{fig:cet} depicts CET (Original). Overall, we observe that standard CET and the one utilizing original labels perform better than distillation. However, for entropy values greater than $0.8$ distillation-based consistency objective results in a model with better generalization on the activity recognition task. This ablation demonstrates that enforcing consistency over exits on clean and perturbed examples is largely sufficient to create a robust multi-exit model.

\begin{figure}[!htbp]
\centering
\subfloat[HHAR]{\includegraphics[width=0.5\textwidth]{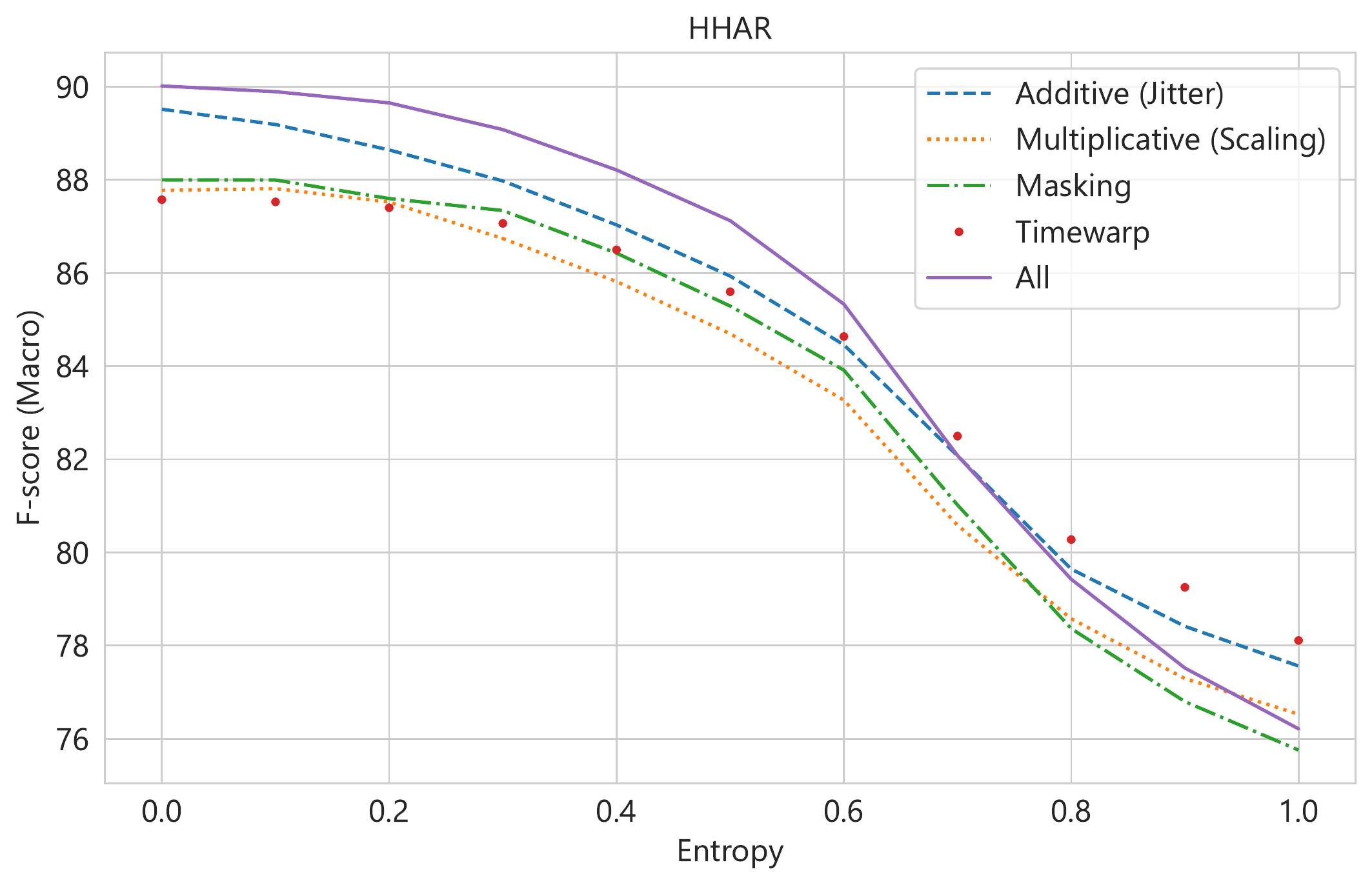}}
\subfloat[SleepEDF]{\includegraphics[width=0.5\textwidth]{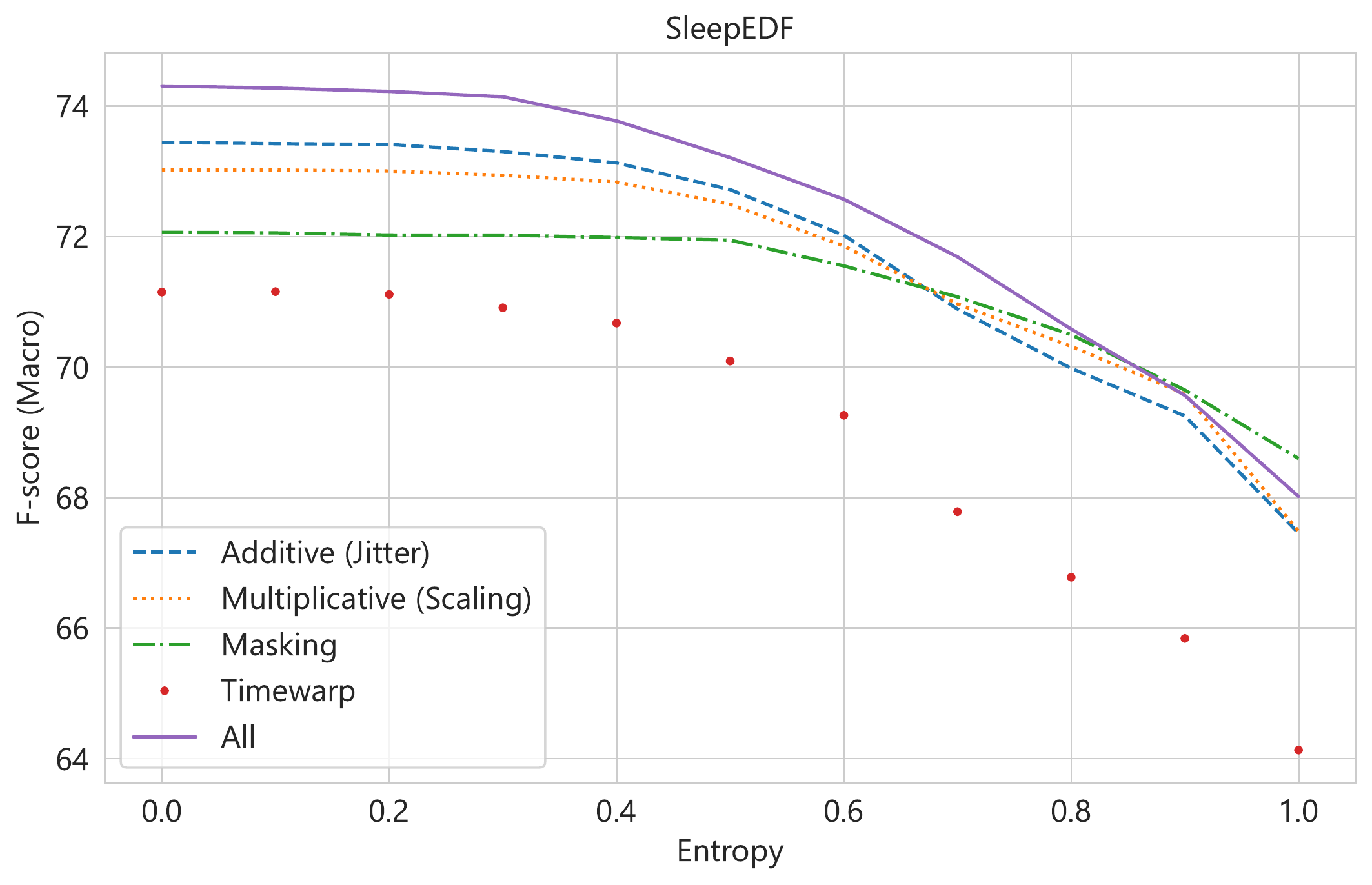}}
\caption{Comparison of different augmentations or input perturbations on the effectiveness of consistency training.}
\label{fig:aug_abl}
\end{figure}

\begin{figure}[!htbp]
\centering
\subfloat[HHAR]{\includegraphics[width=0.48\textwidth]{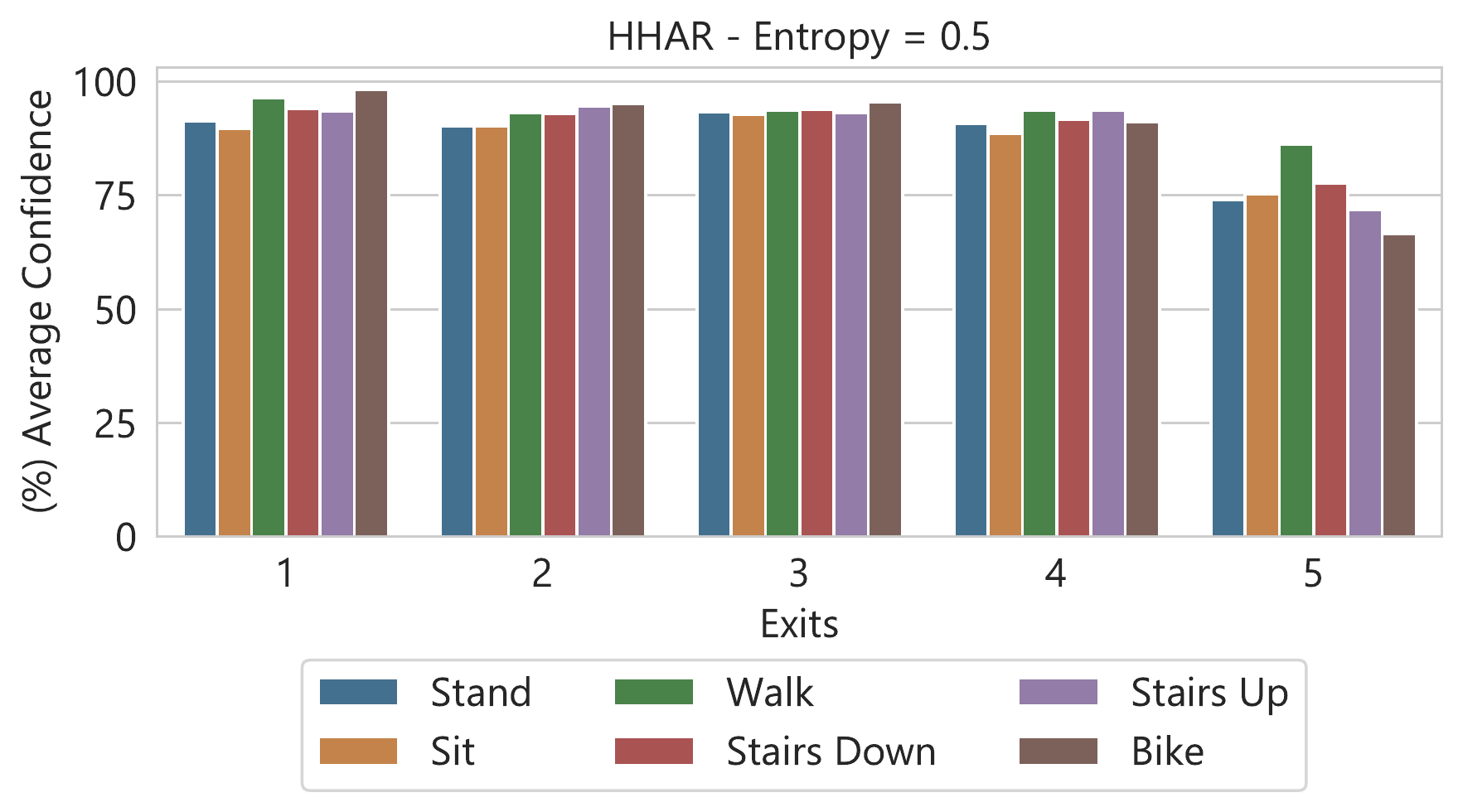}} 
\subfloat[SleepEDF]{\includegraphics[width=0.51\textwidth]{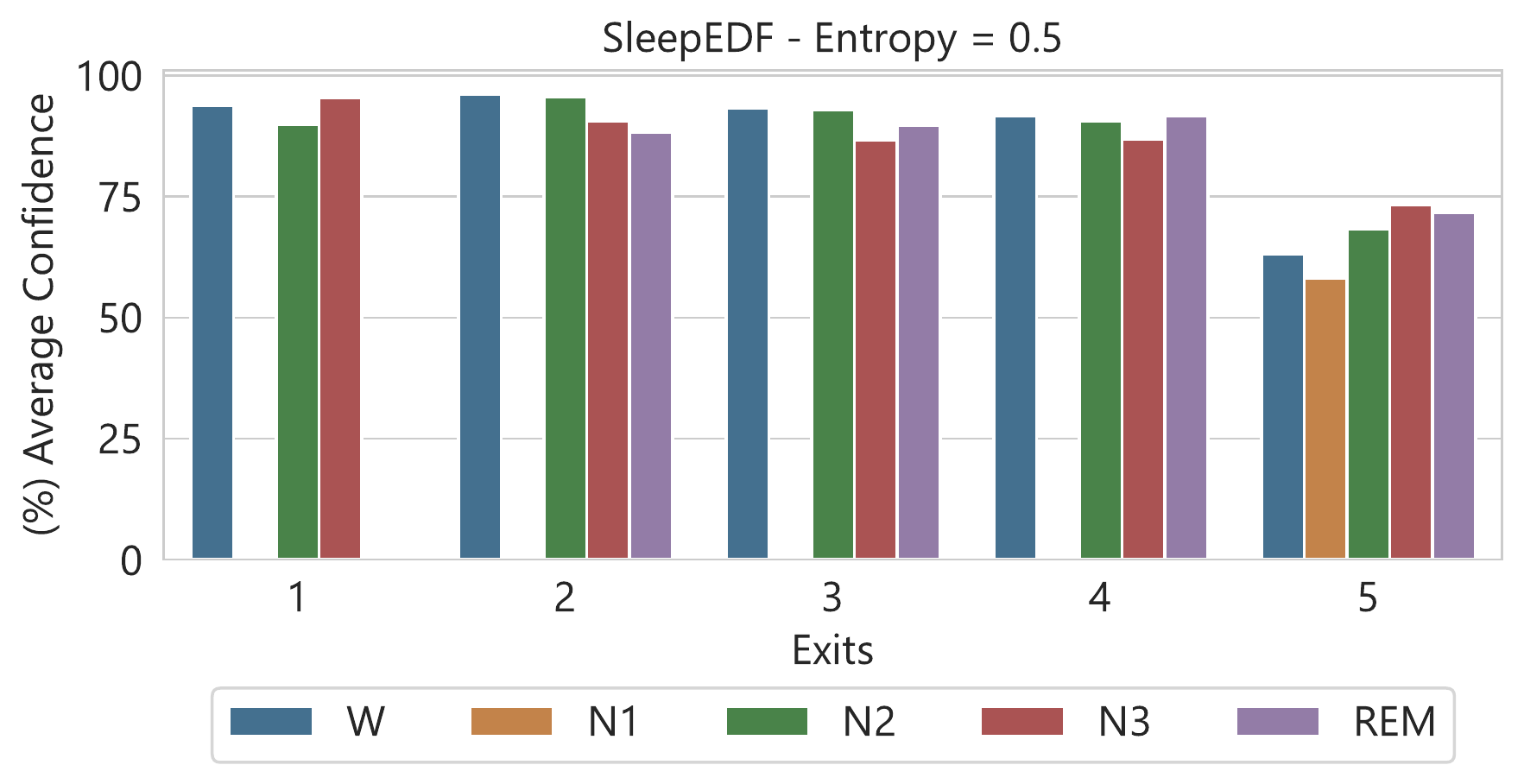}}
\caption{Comparing average confidence levels (i.e., softmax scores) per exit for each class label of CET based multi-exit model.}
\label{fig:avg_confidence}
\end{figure}

In order to understand the effect of applying different perturbations or augmentations on consistency objectives, we conduct an ablation to train a model with each of the specified transformation functions. We compare the performance by utilizing all the perturbations across a range of entropy values as earlier. From Figure~\ref{fig:aug_abl}, we notice that additive noise (or jitter) and masking perturbations are important as compared to time-warping signals. However, combining them results in an overall better generalization across the considered learning tasks. Furthermore, Figure~\ref{fig:avg_confidence} shows the proportion of instances in our test set that are predicted with high confidence (i.e., higher softmax score) using a multi-exit model trained with CET strategy for entropy threshold of $0.5$. We first observe that across all considered datasets, instances with different labels are not predicted with the same level of confidence. For example, the $N1$ class in the sleep stage scoring task is always deferred to the last exit, hinting at the hardness of the example as it can not be predicted with earlier exits with certainty within the specified entropy. Similarly, we note that the average confidence of instances exiting from the last exit has lower confidence from the model, which further indicates that difficult to classify examples to reach the end. Finally, on the considered learning task, per exit confidence for each label was found to be largely consistent, indicating strong generalization of each exits' classifier. 

\section{Conclusion and Future Work}
\label{sec:conclusion}
We propose consistent exit training (CET), a novel framework to train multi-exit architectures for sensory data and improve their quality-efficiency trade-offs. Our conceptually simple and architecture-agnostic approach enforces the prediction invariance over different input perturbations with a consistency objective to make a network to produce similar predictions on clean as well as a perturbed version of the inputs. Our technique induces a stronger regularization to enable the multi-exit model to generalize significantly better than naive exit-wise loss and other baselines when encountering increasing entropy thresholds (or hard to classify examples) while maintaining a similar average exit. Specifically, CET provides large and robust improvements in recognition rate over the existing training procedures for multi-exit models in computationally constrained scenarios. There are a few interesting research directions for future work that remain unexplored in the paper: 1) CET can work seamlessly with unlabeled inputs and does not depend on the availability of ground-truth annotations as labels are acquired from the model output on clean examples. This intuitive property allows our framework to exploit large-scale datasets without class labels, which are inexpensive to acquire. It would be interesting to study CET by taking advantage of perturbations for training MEAs in a semi-supervised way, 2) in this work, we opt for entropy threshold to decide on the early exit of the instance. Other criteria that can take battery or resource availability of the device can be incorporated and studied to further improve inference efficiency, 3) as our approach is agnostic to neural architecture, it can be combined in a straightforward manner with neural architecture search methods to further improve the design of multi-exit models, and 4) our approach is orthogonal to compression and quantization schemes for improving neural network efficiency, in future work we aim to study the fusion of multi-exit models with network sparsification techniques.

\section*{Acknowledgements}
The author would like to thank Johan Lukkien for the valuable feedback and help with this work.

\bibliographystyle{ACM-Reference-Format}
\bibliography{main}
\end{document}